\documentclass[review]{elsarticle}

\usepackage{lineno,hyperref}
\modulolinenumbers[5]
\usepackage{graphicx}
\usepackage{adjustbox}
\usepackage{lipsum}
\usepackage{subcaption}
\usepackage{amsmath}
\usepackage{algorithm}
\usepackage{float}
\usepackage{algorithmic}
\usepackage{tabu}
\usepackage{lscape}
\usepackage{longtable}
\usepackage[table,xcdraw]{xcolor}
\newcolumntype{C}[1]{>{\centering\let\newline\\\arraybackslash\hspace{0pt}}m{#1}}
\usepackage{booktabs}
\usepackage{tabularx}
\usepackage{rotating}
\journal{Journal of Neurocomputing}

\begin{document}

\begin{frontmatter}

\title{A scene perception system for visually impaired based on object detection and classification using multi-modal DCNN}

\author{Baljit Kaur, Jhilik Bhattacharya}
\address{Thapar Institute of Engineering and Technology, Patiala}




\begin{abstract}
This paper represents a cost-effective scene perception system aimed towards visually impaired individual. We use an odroid system integrated with an USB camera and USB laser that can be attached on the chest. The system classifies the detected objects along with its distance from the user and provides a voice output.  Experimental results provided in this paper use outdoor traffic scenes. The object detection and classification framework exploits a multi-modal fusion based faster RCNN using motion, sharpening and blurring filters for efficient feature representation.
\end{abstract}


\end{frontmatter}


\section{Introduction}

\begin{figure}
  \centering
  \includegraphics[scale=.75]{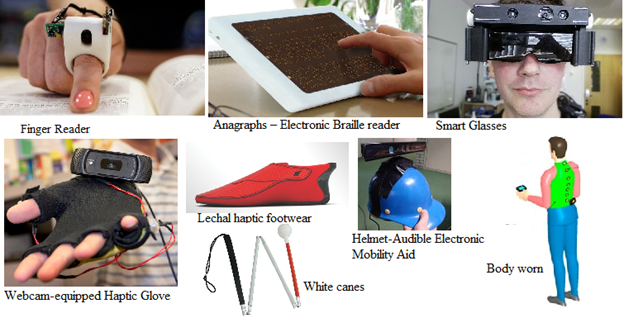}\\
  \caption{A few assistive technological devices for blind aid}\label{1}
\end{figure}

\begin{figure}
  \centering
  \includegraphics[scale=.75]{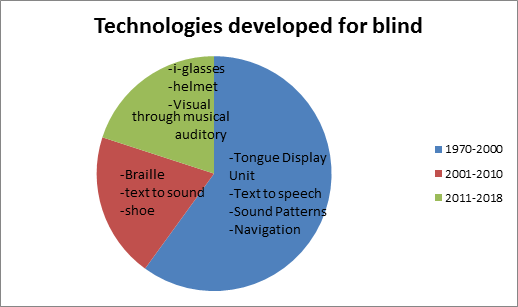}\\
  \caption{Time span representation of usage of particular technologies}\label{2}
\end{figure}

Navigation of blind people is an important issue to be considered. They most commonly use white canes for obstacle detection, meanwhile memorizing all locations they are getting familiar with. In a new or unfamiliar environment they totally depend on individuals passing by to enquire for a certain area. In the world of sophisticated technology along with various sensors, there should be a system with the most basic innovation to make their life a bit easier. This innovation should complement the white canes: give alarm to the user about obstacles a few meters away and give direction for going to a particular area. The navigational aid should also be provided using their other, but now stronger senses like hearing, touch, smell etc. Traditional navigation aid methods such as white canes are very limited and are unable to provide a complete scene perception. For example a white cane may just provide information about the presence or absence of an obstacle. However no information about the kind of obstacle is available. In many cases it may be important to know the obstacle type; specifically if it is a door needed to be opened or a stair required to be climbed. Even information about moving objects and their direction of motion is important. The utilization of computer framework advances for navigational help solutions is relatively recent, for example a smartphone based navigation system(ARIANNA) for both indoor and outdoor environment is available \cite{croce2014arianna}. Work contributed by different researchers in this area can be discussed in terms of sensors used for input; output representation type; and hardware gear, used for either or both. Commonly the scene is perceived using ultrasonic sensors\cite{kay2000auditory} or by extracting images/videos using vision sensors\cite{bach2003sensory}. The later is discussed in more detail in Section \ref{LR}. The output of the above systems(provided in Table \ref{tab1}) can be in the form of tactile image\cite{bach2003sensory}; tongue display through voltage pulse\cite{kaczmarek2011tongue}; sound patterns/musical auditory information\cite{auvray2007learning}\cite{abboud2014eyemusic} etc. Common wearable helping aids for blind include:

 \begin{enumerate}
   \item Wearable tactile harness-vest display to give instructions about directional navigation using six vibrating motors\cite{jones2006tactile}.
   \item A belt associated to a computer along with ultrasonic sensors gives acoustic response in guidance mode, where the system knows about the target and user is guided using tactile signal; image mode, the user is demonstrated about the environment using tactile image. It translates visual of the scene into tactile or acoustic information to permit secure and fast walk\cite{shoval1993navbelt}.
   \item Helmet mounted with ultrasonic chirps and speakers. It amplifies echoes produced by ultrasonic sounds for locating objects in space\cite{sohl2015device}.
   \item Ultrasonic smart glasses work with ultrasonic waves to detect obstacle\cite{agarwal2017low} and many more have been developed for blind aid.
 \end{enumerate}
 Some of these are shown in Figure \ref{1}. A graphical representation showing the distribution of type of work done for blind during various time spans is given in Figure \ref{2}. It is observed that although scene perception via TDU, tactile images, sound patterns are explored and used since $1970's$, the last decade particularly witnesses the marketization of a lot of wearable devices.\\
Alongside developing different blind-aid systems there is also a thorough investigation about suitability and acceptability of these systems. For example,it is argued \cite{gori2016devices} that there are many problems that can be faced with these devices, such as (i)invasive that are coving ears, blocking the tongue, involve use of hands etc. So with these, visually impaired people can never feel free. They will have to hold the device whose size and weight is too big and high. Moreover, these are easy to handle by children. (ii)The user can feel cognitive load as these devices may require lot of attention, that can causes distraction from the primary task. (iii)These devices require lot of training for their usage which is difficult especially for children. (iv)These devices can even perform unsatisfactorily compared to the overhead of using them. (v)The cost of the devices are too high to afford by a common person. (vi)Lastly many of these devices are still at their prototype stage or are tested at pre-clinical level only and they are not available for desired task.
In this paper, we particularly focus on scene perception via image processing algorithms. We are integrating a low-cost, light weight, simple, easily wearable system that will help detect and classify objects on the way of user along with their distances. In this context Section \ref{LR} reviews work particularly focused on object and obstacle detection, navigation assistance using images. Section \ref{S3} discuss the proposed system in detail. Due to wide application and enhanced performance of CNNs for object detection and classification, we reviewed different CNN architectures and their suitability for the current task. Experimental results are shown in Section \ref{S4}. Conclusion is presented in Section \ref{S5}.
\begin{landscape}
\begin{longtable}[c]{|l|p{6cm}|p{6cm}|p{5cm}|}
\caption{Blind aid technologies using different sensor inputs}
\label{tab1}\\
\hline
References & Results & Device used & Method \\ \hline
\endfirsthead
\multicolumn{4}{c}%
{{\bfseries Table \thetable\ continued from previous page}} \\
\hline
References & Results & Device used & Method \\ \hline
\endhead
Paul and Stephen(2003)\cite{bach2003sensory} & Voice message about visual scene & Braille & Convert the image from a video camera into a tactile image. \\ \hline
Kaczmarek(2011)\cite{kaczmarek2011tongue} & Creates real-time tactile images on the tongue & Tongue display unit (TDU) & TDU is a programmable pulse generator that delivers dc-balanced voltage pulses suitable for electrotactile stimulation of the anteriordorsal tongue, through a matrix of surface electrodes. \\ \hline
Matsuda et al.(2008)\cite{matsuda2008finger} & Allow communication between deaf-blind persons. & Mechanical fingers used to transmit Braille symbols & Finger Braille recognition system \\ \hline
Jones et al.(2006)\cite{jones2006tactile} & Navigation Assistance & harness-vest & Convert navigation information into tactile inputs. \\ \hline
Shraga et al.(1993)\cite{shoval1993navbelt} & Detect Obstacles & Computer, ultrasonic sensors and stereophonic headphones & The acoustic signals are transmitted as discrete beeps or continuous sounds. \\ \hline
Sohl-Dickstein et al.(2015)\cite{sohl2015device} & Navigation aid and object perception & Ultrasonic chirps & Amplifies echoes produced by ultrasonic sounds to locate objects. \\ \hline
Kay(2000)\cite{kay2000auditory} & Navigation aid & Ultrasonic transmitter and two microphones & Translate echoes in sounds for navigating and scanning objects. \\ \hline
Auvray et al.(2007)\cite{auvray2007learning}& Audio output for visual input & Webcam & The Vibe device converts a video stream into a stereophonic sound stream. \\ \hline
Abboud et al.(2014)\cite{abboud2014eyemusic} & Visual information through a musical auditory experience. & Camera & An algorithm conveys shape, location and color information using sound. \\ \hline
Joselin and Rene(2012)\cite{villanueva2012optical} & Detect walls, openings, and vertical roads & IR, LED and a photodiode & Pulses emitted by LED, retro diffused light detected by the photodiode. \\ \hline
Mun-Cheon et al.(2015)\cite{kang2015novel} & Obstacle(not type) & Glasses-type vision camera & Deformable Grid (set of vertices and edges with neighborhood system), \\ \hline
Van-Nam et al. (2015)\cite{hoang2015obstacle} & moving objects (e.g., people) and static objects (e.g. trash, plant pots, fire extinguisher) and audio warning & Electrode matrix and mobile  Kinect, RF transmitter, & The color image, depth image, and accelerometer information provided by Kinect \\ \hline
Shaomei et al. (2017)\cite{wu2017automatic} & Give voice message for facebook feeds to the blind user. & Facebook & Artificial intelligence \\ \hline
Shweta et al. (2018)\cite{jaiswal2018small} & Helps to avoid obstacles and give its approximate distance & Ultrasonic sensors & Ultrasonic signals \\ \hline
Rohit et al. (2017)\cite{agarwal2017low} & Indication of obstacle with distance \textless=300cm using buzzer & Ultrasonic Smart glasses & Ultrasonic waves. \\ \hline
Robert et al. (2018)\cite{katzschmann2018safe} & Detect obstacle with distance & Sonar belt and infrared time-of-flight distance sensors & Infrared light \\ \hline
\end{longtable}
\end{landscape}

\section{Literature Review} \label{LR}

As vision is an extremely vital sensory system in humans, its loss affects the performance of most activities of daily living; thereby haltering an individual’s quality of life, personal relationships, general lifestyle and career. With the advent of technology, scientists are trying to develop systems to make visually impaired individuals more independent and aware of their surroundings.  It is often helpful to know the scene around you and then have the knowledge about the obstacles. There are a few devices which provide scene perception for example EyeCane and Eye Music. These use infrared ray to translate color, shapes,location and other information of the object/scene into soundscapes(auditory or tactile cues) which the brain can interpret visually \cite{nau2015use}. Use of vision sensors particularly for this purpose is limited. In general, vision sensor input is utilized by visually impaired individuals for reading documents (via OCR) or identifying street signs, hoardings etc (scene segmentation, followed by OCR) \cite{gori2016devices}. Social interaction assistance for individual with visual disability are also provided to some extent in the form of person recognition, facial expression recognition \cite{panchanathan2016social}. A number of devices are developed for blind people to provide them information about presence of obstacles, types of obstacle, their distance etc. This information is further utilized to assist blind people for navigating safely in indoor and outdoor environments. Ruxandra et al. proposed a smartphone based system that indicates the type of obstacle and categorizes it as urgent or normal depending on its distance from the user. Obstacle candidates are tracked with multiscale Lucas - Kanade using SIFT/SURF interest points and urgency of obstacles is identified by motion estimation using homomorphic transforms and agglomerative clustering. Then the image patches of detected obstacle regions are classified with SVM using a visual codebook generated with k-means clustered HOG features. \cite{tapu2013smartphone}. Mun-Cheon et al. indicates  presence of obstacle using deformable grid(DG) which use the motion information accumulated over several frames. DG consists of a set of vertices and edges with n-neighborhood system. This method is accurate as well as robust to the motion tracking error and ego-motion of the camera. The above method detects the object having risk of collision by using the extent of contortion of the deformable grid. However, this method is unable to perform in areas having walls and doors\cite{kang2015novel}. To overcome the issue of detecting walls and doors, Van-Nam et al. developed an electrode matrix and mobile Kinect based obstacle detection and warning system for visually impaired individual in which they detect moving and stationary obstacles(using color and depth information given by Kinect) and warn the user with the help of Tongue Display Unit. The degree of warning depending on the depth information is provided by changing the level of electric signal on the electrode matrix \cite{hoang2015obstacle}. To provide stress free environment to blind people, work is done to detect potholes and uneven surfaces. Aravinda et al. use vision based system along with Laser patterns for detecting potholes and Uneven Surfaces \cite{rao2016vision}. They use Hough transform to detect lines recorded using laser.
 Kanwal et al. provide wall-like obstacle information (through voice message) using Kinect both as camera and depth estimator. Camera detects corners of the obstacle using Harris and Stephens corner detector and its infrared sensor give corresponding depth value for indoor environment \cite{kanwal2015navigation}. Aladren et al. proposed visual and range information based Navigation assistance for visually impaired. They used a consumer RGB-D camera, and take advantage of both range and visual information about floor, walls and obstacle for indoor environment. Their system gives voice commands about the obstacle to the user \cite{aladren2016navigation}. They perform segmentation using range data (via RANSAC) which is further enhanced using mean sift filtering on color data. Sarfraz and Rizvi \cite{sarfraz2007indoor} developed navigation assistance for indoor environment providing depth and object type information including presence of human, doors, hallway or corridor, staircases, elevators, moving objects. They developed individual algorithm for each obstacle to be detected using CannyEdgeDetector. They use camera vision input and text-to-speech synthesized output to provide navigation aid. Table \ref{tab2} briefs some notable visually impaired assistive technologies which clearly portrays the general trend of existing systems in terms of sensors used to perceive input from the environment; output type and devices via which information is provided back to the user and feature extraction technique used for object detection and classification. In this particular work, our main contributions are; \begin{itemize}
        \item We develop a scene perception system which will provide information about objects(object detection and classification from images) in the scene and their relative depth(using laser). We focus on making this system low-cost, light weight, simple and easily wearable emphasising that no explicit training is required to use the system. At the same time, we ensure that maximum information about the environment can be retrieved. The system currently works via restricted voice output which implies that the user can select how much information he wants. He can chose to be informed about scene changes at (a)fixed intervals or (b)a single time only (c)when obstacles are too close.
        \item Most of the work discussed above use corner detectors, key point descriptors such as SIFT/SURF, edge detectors, HOG descriptors for obstacle classification. Inspite of their success, they often suffer from the issues like; background has more key points than candidate objects(or obstacles), absence of key points due to low resolution, poor texture and random motion. This work  exploits the power of the state of the art DCNN due to its high performance for object detection and classification. In this course, we review different kinds of architectures which include single column and  multicolumn CNN. Multicolumn architectures mostly use full image as one column and different patches of the image as second column input. We exploit multicolumn architecture by using different features like image edges($CNN-E_{k}$),optical flow($CNN-O$) or scale space($CNN-G_{t}$) representations along with RGB image($CNN-I$). In this paper, we demonstrate 3 multicolumn architectures; $\{CNN-I, CNN-E_{c},CNN-E_{s}, CNN-E_{p}\}$,$\{CNN-I,CNN-O\}$ and $\{CNN-I, CNN-G_{3},CNN-G_{5}\}$.
        \item In general the most common CNN input is 3-channeled intensity image(RGB). Some contributions focus towards enhancing it by including different parametric inputs. For example a 4-channeled CNN input consist of RGB-D. Another kind of architecture may involve a multispectral fusion using 2 different CNNs, i.e. fusing separate convolutional feature map outputs of RGB and D inputs at early or late stages. In this paper we use a multimodal fusion, where we fuse the convolution feature maps of individual columns of the multicolumn CNN using summation and maximum operation. This proves to be beneficial by accommodating multimodal features with minimal computational space and speed as opposed to fusion techniques via concatenation. For example in our case we fuse outputs of convolutional feature maps for the different multicolumn architectures. Three different fusions are used here: $\{conv5\_F_{I},conv5\_F_{E_{k}}\},\{conv5\_F_{I},conv5\_F_{O}\}$  and  \\ $\{conv5\_F_{I}, conv5\_F_{G_{3}}, conv5\_F_{G_{5}}\}$.
      \end{itemize}

\section{System Details} \label{S3}

\begin{figure}
  \centering
  \includegraphics[scale=0.75]{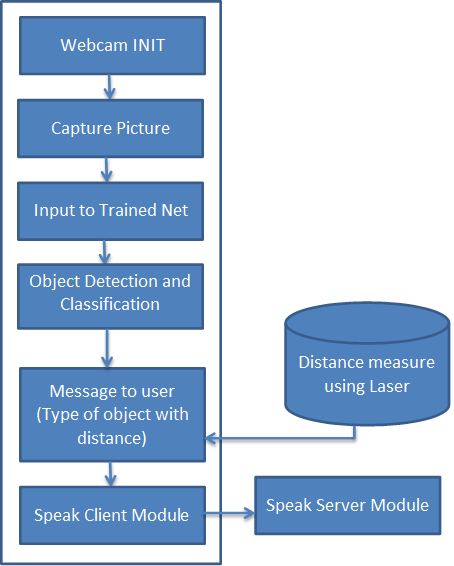}\\
  \caption{Flow chart for the proposed scheme}\label{3}
\end{figure}

The flow chart of system developed is shown in Figure \ref{3}. This system is at its early stage but is already capable to fulfill all of the basic requirements needed:
\begin{enumerate}
  \item Acquisition of a video stream from a webcam with HD resolution (1920x1080, 25 fps)
  \item Detection of multiple objects from the scene, even if the position of object or vehicle is not perfectly in front of the camera.
  \item Detection of one or more candidate objects in the scene.
  \item Generation of a vocal output as a synthesized voice saying the name of the recognized object or vehicle and its distance from the user
\end{enumerate}
\begin{landscape}
\begin{table}
\centering
\caption{Blind aid technologies using vision sensor inputs}
\label{tab2}
\resizebox{\textwidth}{!}{%
\begin{tabular}{|l|p{4cm}|p{4cm}|p{4cm}|}
\hline
References               & Information                                                                                                                                                            & Sensors                             & Image features                                                                                                                                                                                                                \\ \hline
Ruxandra et al.(2013)\cite{tapu2013smartphone}     & Obstacle detection and classification in real time and helps the user in avoiding and recognizing static and dynamic objects like vehicles, pedestrians, bicycles etc. & All sensors available in Smartphone & SIFT, SURF and HOG features extracted. Used SVM classifier for classification                                                                                                                                                 \\ \hline
Aravinda et al.(2016)\cite{rao2016vision}    & Detect potholes and Uneven Surfaces                                                                                                                                    & Laser and  monocular camera         & Hough transform is used to detect the laser projected lines. The intersecting points obtained from the Hough transform are binned to create a histogram for each frame. This feature is dubbed as Histogram of Intersections. \\ \hline
Kanwal et al.(2015)\cite{kanwal2015navigation}      & Detect Wall-like obstacles (pillar, people) and their depth                                                                                                            & kinect                              & Harris \& Stephens corner detection                                                                                                                                                                                           \\ \hline
Alarden et al.(2016)\cite{aladren2016navigation}    & Provide Obstacle free path                                                                                                                                             & RGB-D camera                        & Canny edge detector followed by the probabilistic Hough line transform                                                                                                                                                        \\ \hline
Sarfraz and Rizvi (2007)\cite{sarfraz2007indoor} & Detect human presence, doors, hallway/corridor, staircases, elevators, moving objects in indoor settings.                                                              & Monocular vision camera             & Canny Edge Detector                                                                                                                                                                                                           \\ \hline
Saranya et al. (2018)\cite{saranyareal}    & Detect object                                                                                                                                                          & camera                              & Histogram features of intensity and gradient and Edge linking features.                                                                                                                                                       \\ \hline
\end{tabular}%
}
\end{table}
\end{landscape}
\subsection{Single-board PC: Odroid}
\begin{figure}
  \centering
  \includegraphics[scale=0.70]{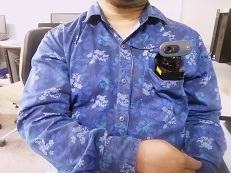}\\
  \caption{Camera and laser attached to the chest}\label{4}
\end{figure}

\begin{figure}
  \centering
  \includegraphics[scale=0.70]{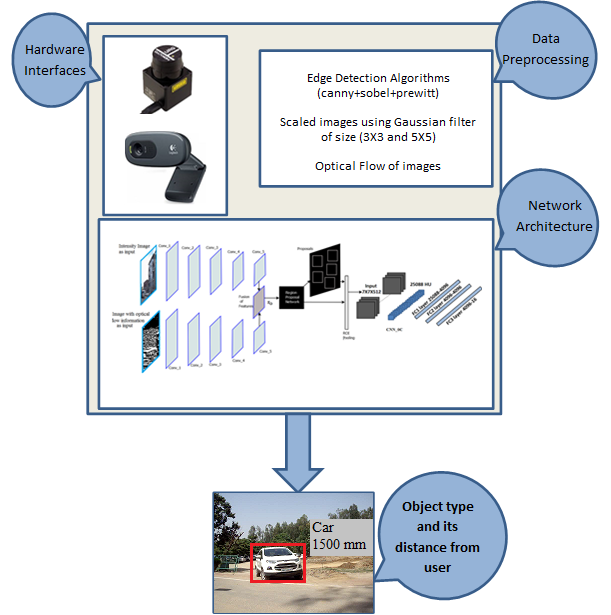}\\
  \caption{System architecture}\label{5}
\end{figure}

Single-board PC is used for testing purpose of the work done.  It is potable and easy to wear. There are several single board computers in the market, among which the most renowned are probably BeegleBone, Raspberry, Odroid, Udoo, Lattepanda; they are released at an impressive cadence in progressively more powerful versions.  We decide to use an Odroid XU4 board to develop our prototype due to the reason that the Odroid XU4 currently is ranked 5th. It is noted that this ranking is not only related to system performance, but also to other aspects that we do not consider mandatory for the present project, such as the platform cost, the availability of software in the web or the presence of communities that support software updates and forums. The main feature which encourage us to opt for Odroid XU4 platform is the processor: the Odroid XU4 has a Samsung Exynos5422 octa-core working at 2 GHz and perform better as compared to other competitors. In addition to that, the presence of an ARM Mali - T628 GPU may be very useful to further improve the processing throughput by exploiting parallel computation, due to the support of OpenCL. Another important aspect is the large amount of available memory, 2 GB of embedded DDR3, which let the image data and its features to get stored easily. Moreover, the memory is extendible up to 64 GB in eMMC format, allowing much faster access as compared to a common SD memory. Finally, two USB 3.0 ports allow us to easily capture high-resolution video streams from two cameras, while only slower USB 2.0 ports are available on the Raspberry. Hokuyo URG-04LX-UG01 scanning laser is attached to the USB to get the distance of the detected object from the user. It is small, affordable and accurate laser scanner and is able to report ranges from $20 mm$ to $5600mm$ in a 240 degree arc with 0.36 degree angular resolution.  Its power consumption, 5V, allows it to be used on battery operated platforms. The Logitech C270 HD high quality Webcam is attached. Its 3 MP camera helps in recording superior quality with good clarity visuals in both day and night time environment, As this webcam has also the feature of effective light correction. Laser and camera can be attached to the chest as shown in Figure \ref{4}.
Odroid is a small PC, capable of running a full Linux distribution. We choose Ubuntu Mint $16.04$ to have access to the well populated and mature community of Ubuntu users and software; consequently, the setup time may be significantly reduced and the procedures needed for having a ready-to-use system are quite the same as those for the x86 PC. The system currently runs at about 1 fps: further experiments with the User Group and appropriate optimization are in progress. The system architecture is shown in Figure \ref{5}.

\subsection{Features}\label{featur}
Features with deep learning for object detection\cite{chu2018deep}, scene perception are widely used in recent research. Among CNNs for object detection, R-CNN, fast RCNN, faster RCNN are widely adapted. R-CNN extract region proposals, compute CNN features and classify the objects. To improve computation ability, Fast R-CNN use region of interest pooling by sharing the forward pass of CNN. These region proposals are created using selective search which is replaced by RPN in faster R-CNN \cite{ren2017faster}. Here a single network composed of region proposal and Fast R-CNN is used by sharing their convolutional features. An option to add segmentation properties to Fast RCNN is enabled by putting an object mask predicting feature with the already occurring branch for bounding box recognition \cite{he2017mask}. It is noticed that these object detection networks fine tuned VGG16 with PASCAL dataset. It is argued VGG16 performs better than AlexNet and GoogleNet. Another widely used network YOLO \cite{redmon2016yolo9000} compose of entirely convolutional layers trained and tested on PASCAL VOC and COCO datasets is quite accurate and fast.

For enhanced performance researchers particularly focus on fine tuning including temporal information and architectural enhancement. \cite{yao2017coupled}, \cite{zhuo2017vehicle} and \cite{wang2016vehicle}fine tuned different pretrained(on ILSVRC-2012 dataset) networks like GoogleNet, Alexnet, Fast-RCNN using their own datasets for vehicle detection. \cite{li2017attentive} \cite{kang2017optimizing} reduce false alarms by introducing context based CNN model or propagated motion information to adjacent frames. Architectural enhancement is done in different ways such as RGB, Depth \cite{aladren2016navigation}\cite{hou2018object}, optical flow \cite{sarkar2017deep} information is treated as different channels and fed to single column CNN or mutilcolumns CNN is trained using different data(global and local patches). Some architectural modification over alexnet is used in \cite{wang2016multi} to \cite{wang2016brain}. For example in \cite{wang2016multi}, the 5th convolutional layer of AlexNet is replaced with a set of seven convolutional layers(referring to 7 different objects), which are piled in parallel with mean pooling  and then fed to the fully connected layers. This network is trained in 2 phases; first individually seven networks are trained for each scene category and secondly their weights are used for the parallel layers and entire network is retrained. This enables classification of scene consisting of these objects more accurately. \cite{lu2015deep} propose a single column CNN having four convolution layers and three fully connected layers, with the last layer giving a softmax probability as output using information extracted from multiple image patches. In this an image is divided into multiple patches, each of the patches are fed to the CNN and from the last layer, feature output is extracted. For the output of all the image patches, some statistical aggregation such as minimum, maximum, median and averaging of all these features are performed and the final softmax is taken as the output.  \cite{lu2015rating} train a 2 column CNN in which the input for the network is considered as global image as well as local image patches. \cite{wang2016brain} propose multicolumn CNN model in which CNN (trained with style attribute prediction to predict different style attributes for input) treated as additional CNN column is then added to the parallel input pathways of a global image column and a local patch column.\\
Incase of multicolumn  CNN architectures, the feature output of the different columns are fused at various stages. \cite{liu2016multispectral} demonstrate the use of a multispectral CNN where different kind of images like intensity image, thermal image etc are used for training and obtained feature maps are fused using concatenation.
In this work we use edge, optical flow, scale space representations along with RGB intensity images and fuse the convolutional feature maps before applying RPN. We use edges obtained from canny, sobel and prewitt edge detection algorithms; scales values of $t=3$ and $5$; orientation data of optical flow for extracting pertinent features through different scale and orientation information of an object. The PASCAL VOC dataset is used for extracting the features from 5th convolutional layer $(conv5)$ of VGG 16. The steps of training and testing are shown in Algorithm \ref{algo1}. Networks used for intensity image, edge image (canny, sobel, pewitt), Gaussian image, optical flow image are named as ${CNN-I, CNN-E(CNN-E_{C},CNN-E_{S},CNN-E_{P})}$, ${CNN-G(CNN-G_{3}, CNN-G_{5})}$ and $CNN-O$ respectively. Features from $conv5$ of $CNN-I$ referred as $conv5\_F_{I}$ are fused with features of $conv5$ of $(CNN-E_{C} (conv5\_F_{Ec}), CNN-E_{S} (conv5\_F_{Es}), CNN-E_{P} (conv5\_F_{Ep}))$ separately for resulting features map $(F_{E})$. In the same manner, feature map $(F_{O})$ is obtained by fusing $conv5\_F_{I}$ and $conv5\_F_{O}$.  Feature map $(F_{G})$ is obtained by fusing $conv5\_F_{I}, conv5\_F_{G3}$ and $conv5\_F_{G5}$. These feature maps are further passed for ROI pooling and classified using two different classification networks $CNN\_1C$ and $CNN\_0C$ as shown in Figure \ref{E}. $CNN\_0C$ has three fully connected layers while $CNN\_1C$ has one convolutional layer with three fully connected layers.

\subsection{Fusion}
A common way of fusing features is by concatenating them\cite{liu2016multispectral}. However this will lead to the enlarged size of feature map which take a lot of computational time and space. So in this paper, we used addition and/or maximum of features which will retain the size of feature map. There are three cases of fusion of feature extraction:
\begin{enumerate}
  \item In first case of edges, features are fused by adding the features ($conv5\_F_{I}$ and  $conv5\_F_{Ec}$), ($conv5\_F_{I}$ and $conv5\_F_{Es}$) and ($conv5\_F_{I}$ and $conv5\_F_{Ep}$). Further taking the maximum of these three gives the final feature map $(F_{E})$ as shown in equation \ref{eqE}. The process using different network classifiers ($CNN\_1C$ and $CNN\_0C$) is shown in Figure \ref{E}.
      \begin{equation}\label{eqE}
        F_{E}=max \begin{cases}
        \text{$conv5\_F_{I}+conv5\_F_{Ec}$}\\
        \text{$conv5\_F_{I}+conv5\_F_{Es}$}\\
        \text{$conv5\_F_{I}+conv5\_F_{Ep}$}
    \end{cases}
      \end{equation}
\begin{figure}
  \centering
  \includegraphics[width=\textwidth]{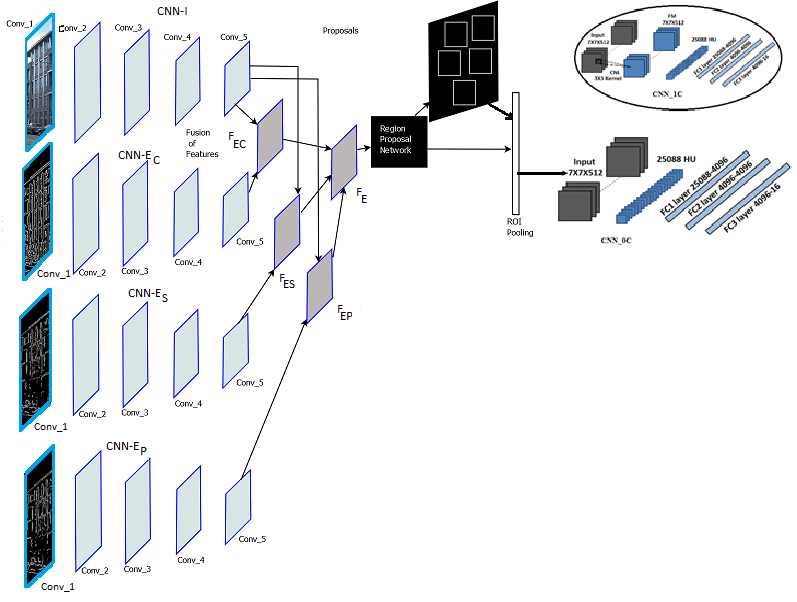}\\
  \caption{Multimodal object detection and classification using RGB and edge features}\label{E}
\end{figure}

  \item In case of optical flow, $Conv5\_F_{I}$ are fused (added) to orientation features $(conv5\_F_{O})$. Feature map $F_{O}$ is obtained as shown in equation \ref{eqO}. The whole process is shown in Figure \ref{O} with classifier network $CNN\_0C$. It is also done with $CNN\_1C$.
      \begin{equation}\label{eqO}
        F_{O}=conv5\_F_{I}+conv5\_F_{o}
      \end{equation}
\begin{figure}
  \centering
  \includegraphics[width=\textwidth]{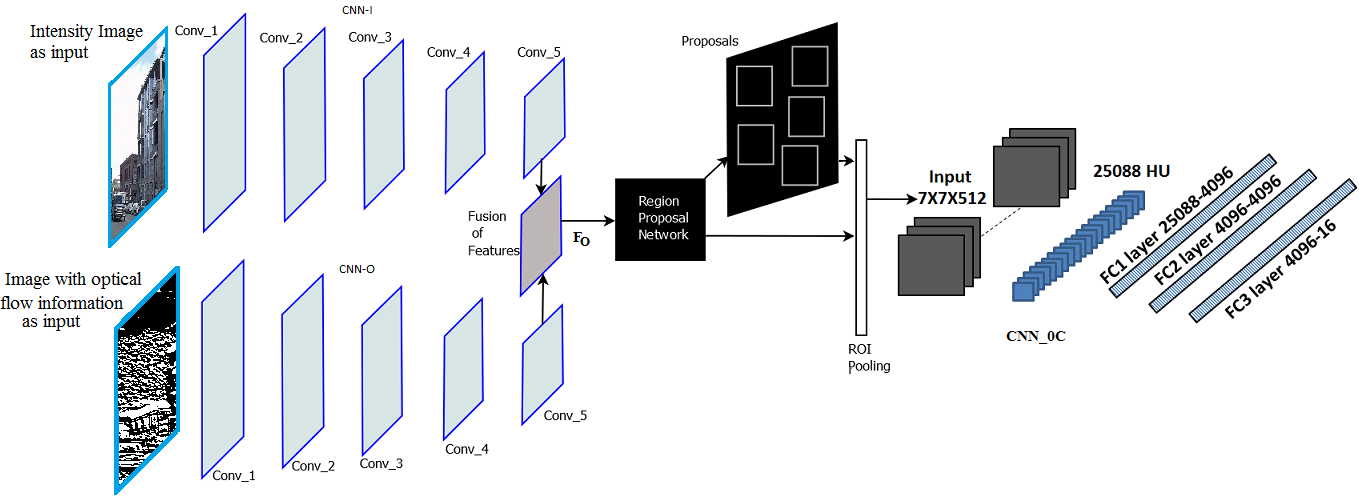}\\
  \caption{Multimodal object detection and classification using RGB and optical flow features}\label{O}
\end{figure}
  \item For scaled images, the fusion is done by taking maximum of the features of $conv5\_F_{I} , conv5\_F_{G3}$ and $conv5\_F_{G5}$. The feature map $(F_{G})$ is obtained using equation \ref{eqG}. The process is represented in Figure \ref{G}.
      \begin{equation}\label{eqG}
        F_{G}=max \begin{cases}
        \text{$conv5\_F_{I}$}\\
        \text{$conv5\_F_{G3}$}\\
        \text{$conv5\_F_{G5}$}
    \end{cases}
      \end{equation}
\begin{figure}
  \centering
  \includegraphics[width=\textwidth]{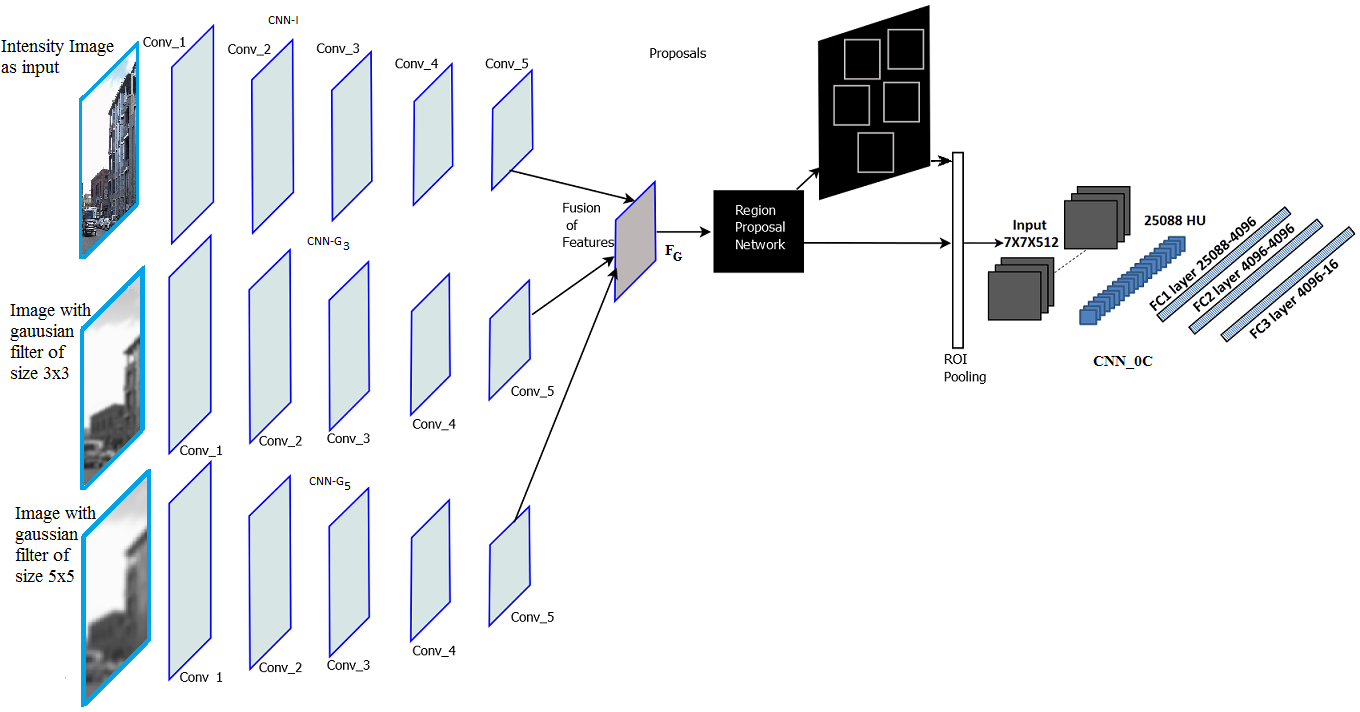}\\
  \caption{Multimodal object detection and classification using RGB and Gaussian scaled features}\label{G}
\end{figure}
\end{enumerate}

\subsection{Depth Data}
During testing, the trained network will detect and classify the object or vehicle coming on the way of user and laser will tell the distance of the detected object or vehicle. The distance factor is added to the system by mapping the laser data with the vehicle detection and classification. The laser gives the data in the form of polar value $(p)$, distance$(z)$ and angle$(a)$ of the object from the center point of laser. From the polar value, Cartesian coordinates($x$, $y$) of the object are calculated using equation \ref{Leq}.
\begin{equation}\label{Leq}
  x=\rho*\cos\alpha, y=\rho*\sin\alpha
\end{equation}

So the laser data is represented in the form of ($x, y, z$). At the same time, the image is captured by the camera, the pixel value of the particular object from that image is acquired in the form of $(q, r)$. Both the captured scenes from camera and laser sensor, are divided into grids for the mapping purpose. Some of the instances of the process of getting data is shown in Figure 9. The grid $(G_{c})$ containing pixel value of an object is mapped with grid $(G_{l})$ of laser containing the object. A mapping from $(q, r,G_{c})$ to $(G_{l})$ is accomplished by using a function represented in equation \ref{Leq1} from which the distance($z$) of the object is figured out.
\begin{equation}\label{Leq1}
  f:A_{i}\rightarrow B_{i}
\end{equation}
where$A$ represents Data of camera $(q, r,G_{c})$ and $B$ represents output $(G_{l}, z)$ and $i$ represents the index.
\begin{figure}
  \centering
    \begin{subfigure}[t]{0.5\textwidth}
        \centering
        \includegraphics[width=\textwidth]{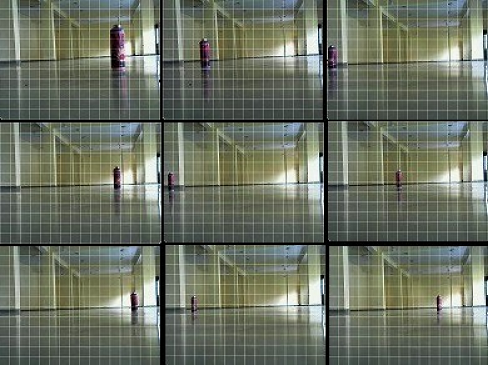}\\
        \caption{Instances of images(captured via camera) in grid style}\label{pl1}
    \end{subfigure}
    ~
    \begin{subfigure}[t]{0.5\textwidth}
        \centering
        \includegraphics[width=\textwidth]{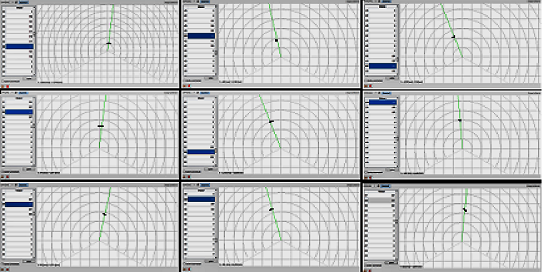}\\
          \caption{Instances of images(generated via laser) in grid style}\label{pl2}
    \end{subfigure}
    \caption{Mapping of laser data with images capture by camera} \label{pl}
\end{figure}

\begin{algorithm}[H]
\caption{Obstacle predictor and classifier along with distance for blind}\label{algo1}
\textbf{Input:} {Training Set P =$\{(x_{1},y_{1}),\cdot\cdot\cdot,(x_{n}, y_{n})\}$,$x_{i} \epsilon X$, number of classes C, $y_{i} \epsilon  Y=y_{1},y_{2},\cdot\cdot\cdot,y_{c}$}\\
\textbf{Output:} {$R= [O \epsilon X, y \epsilon  Y, Z ]$, O is object detected, y is its label and Z is distance of object from user.} \\
1: Divide P into 3 equal parts $P_{j}$ where j=1,2,3.\\
2: For j=1 to 3\\
       \hspace*{3ex}  (i)  Extract edges, scale space and optical features of intensity images(I) of \hspace*{6ex}set $P_{j}$.\\
       \hspace*{3ex} (ii) Fuse convolutional feature maps of I with A, where  $A \epsilon \{E,G,O\}$ \hspace*{6ex}in which E is for edges(canny($E_{c}$), sobel($E_{s}$), prewitt($E_{p}$)), G is for \hspace*{6ex}Gaussian($G_{3}$,$G_{5}$),O is for optical flow.\\
       \hspace*{3ex} (iii)Pass the fused feature maps for ROI pooling using Region Proposal \hspace*{6ex}Algorithm as given in Algorithm \ref{algo2}\\
       \hspace*{3ex} (iv) Create two networks having 1convolution and 3 fully connected \hspace*{6ex}layers$(CNN\_1C)$ and only three fully connected layers$(CNN\_0C)$ using \hspace*{6ex}shared weights of VGG16\\
       \hspace*{3ex} (v)  Train this network with $P_{j} \bigcup \{x_{i}, y_{i}\}$\\
       \hspace*{3ex} (vi) Output the trained net which can predict and classify the obstacle.\\
    EndFor\\
3: Using trained net, features of test set are extracted and SVM is trained to classify.\\
4: Calculate accuracy by comparing the predicted and actual output.
5: Laser is used for extracting the distance Z of obstacle predicted and classified by the net.\\

\end{algorithm}

\begin{algorithm}[H]
\caption{Region Proposal Algorithm}\label{algo2}
$(a)$	The first step is that image is given as input to a convolution network which will output a set of convolutional feature maps on the last convolutional layer\\
                    $(b)$	Then a sliding window of size n x n is run spatially on these feature maps. A set of anchors are generated which all have the same center but with different aspect ratios and different scales. All these coordinates are computed with respect to the original image.\\
                    $(c)$	For each of these anchors, a value p* is computed as shown in equation \ref{eqP*} which indicated how much these anchors overlap with the ground-truth(GT) bounding boxes.\\
                   \begin{equation}\label{eqP*}
                     P^{*}=max \begin{cases}
                   1 \text{\hspace*{3ex}if $IU>0.7$}\\
                   -1 \text{\hspace*{3ex}if $IU<0.3$}\\
                   0 \text{\hspace*{3ex}otherwise}
                    \end{cases}
                  \end{equation}
                  where IU is intersection over union and is defined below in equation \ref{eqnIU}:\\
                  \begin{equation}\label{eqnIU}
                    IU=\frac{Anchor\bigcap GTbox}{Anchor \bigcup GTbox}
                  \end{equation}
\end{algorithm}

\section{Experimental Results}\label{S4}
\begin{figure}
  \centering
  \includegraphics[width=\textwidth]{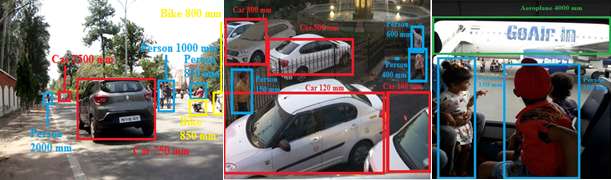}\\
  \caption{Results of object detection and classification along with their distance from the user}\label{res}
\end{figure}

As described in Section \ref{featur}, the features from 5th convolutional layer of VGG16 are extracted using different image representation such as intensity or RGB image, edges(canny, sobel, prewitt), scale space using different gaussian filters$(t=3,5)$. In training phase, network is trained using PASCAL dataset (with these image representations); that is divided into 3 parts ($P1,P2,P3$). The created networks $CNN\_1C$ and $CNN\_0C$ are trained on different datasets ($P1, P2, P3$) individually with learning rate $(0.01 \rightarrow 0.005)$. Average of weights and bias of 3 trained net is calculated for getting the final net in both cases. Further testing with our own dataset is done using these two trained nets.
During testing, a person is blind folded to realistically simulate the experience of visually impaired people. Testing is done at different spots and timings. There can be a situation when no movement is shown by the camera for a long time. The reason for no movement shown by the camera could actually be because of lack of vehicular movement or the person facing a wall or something obstructing the view of the camera. Another hindrance is the shadowy effect of trees and poles during the afternoon, which hampers the quality of videos. During evening, the traffic movement is high extracting data from such obstructed roads is another struggle. Overcoming the issues, different speech messages are prepared. Results of methods proposed by different researchers are presented in table \ref{tabsvm}. Results of the proposed work are presented in the form of (1) Vehicle Detection and Classification using deep neural network. (2) The distance of vehicle from a blind user using laser as shown in Figure \ref{res}.
\begin{table}[]
\centering
\caption{Results of already existing Methods}
\label{tabsvm}
\begin{tabular}{|l|l|l|}
\hline
Method       & DS1 & DS2 \\ \hline
HOG\cite{tapu2013smartphone}          & 50  & 52  \\ \hline
Fast-RCNN\cite{girshick2015fast}  & 54  & 61  \\ \hline
Faster-RCNN\cite{ren2017faster}  & 60  & 66  \\ \hline
Yolo\cite{redmon2016yolo9000}         & 65  & 68  \\ \hline
\end{tabular}
\end{table}

\begin{figure}[H]
  \includegraphics[width=\textwidth]{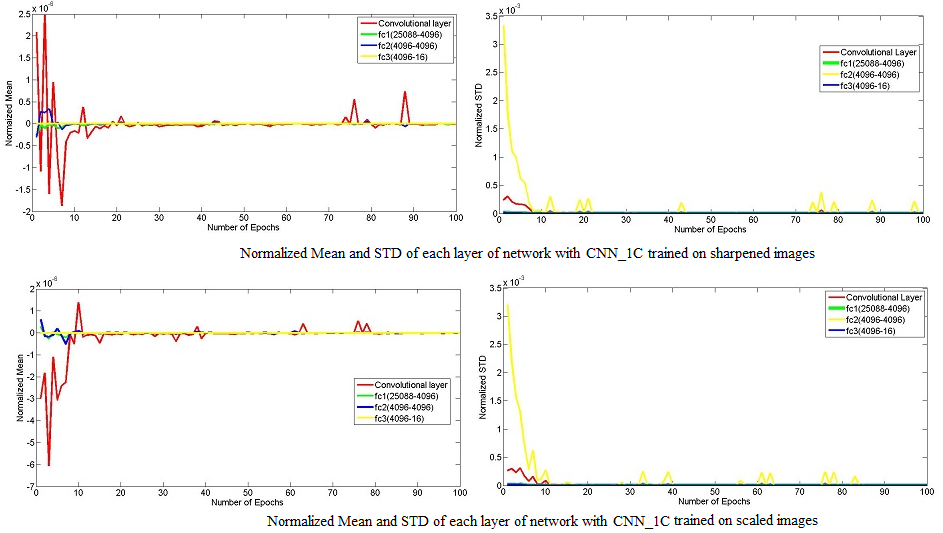}\\
  \caption{The norm of the means and standard deviations of the weights gradients for each layer of network $CNN\_1C$ as function of the number of training epochs. The values are normalized by the L2 norms of the weights for each layer.}\label{cnn1c}
\end{figure}

\begin{figure}[H]
  \includegraphics[width=\textwidth]{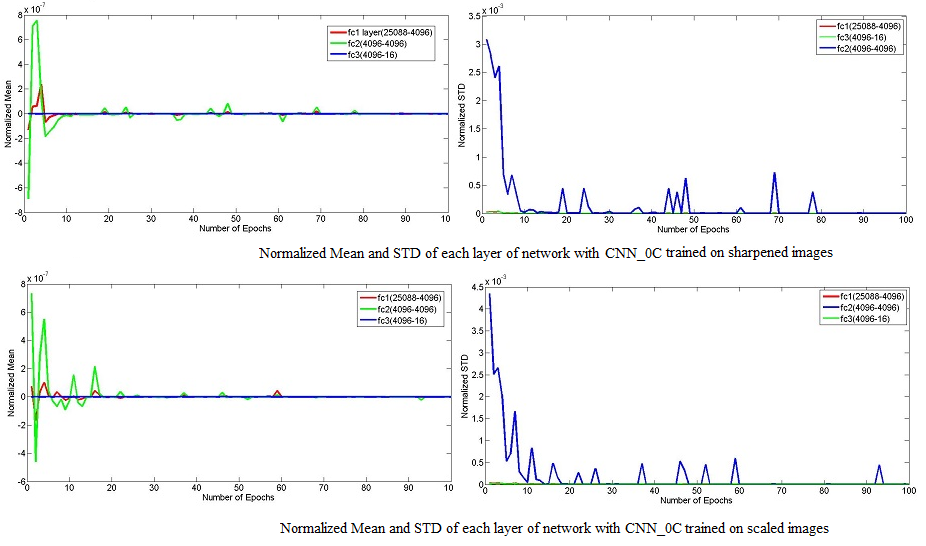}\\
  \caption{The norm of the means and standard deviations of the weights gradients for each layer of network $CNN\_0C$, as function of the number of training epochs. The values are normalized by the L2 norms of the weights for each layer}\label{cnn0c}
\end{figure}

Accuracies for different network architectures in terms of object detection and classification for normal, edges and scaled images are shown in Table \ref{tabres}. Results depict that scaled and edge images give good results as compared to normal and network architecture having 1 convolutional layer and 3 fully connected layers proved to be good rather than only 3 fully connected layers. While comparing both architectures, net trained with normal data show noticeable improvement with $CNN\_1C$ while other are almost similar. This also proves that although using convolutional layer the improvement of result is quite visible when we are using normal data but it is not that visible when we are using some other features. So, using features is almost as beneficial as adding convolutional layer. Moreover, when net trained with edges/scaled/optical flow($OF$) of images are tested with edges/scaled/optical flow of images respectively, accuracy is higher rather than net trained with edges/scaled images are tested with normal images.
\begin{table}[]
\centering
\caption{Accuracies of Different network architectures for edges, scaled and normal images}
\label{tabres}
\begin{tabular}{|l|l|l|l|l|l|l|}
\hline
Learning Rate/testsets & Edges & Gaussian & N-E & N-G  & OF   & Normal \\ \hline
CNN\_0C                & 79    & 77       & 78  & 78.5 & 77.5 & 65     \\ \hline
CNN\_1C                & 81.5  & 82       & 79  & 79.5 & 79.8 & 81     \\ \hline
\end{tabular}
\end{table}
Figure \ref{cnn1c} and \ref{cnn0c} shows the norm of the means and standard deviations of the weights gradients for each layer of network $CNN\_1C$ and $CNN\_0C$ respectively as function of the number of training epochs. The values are normalized by the L2 norms of the weights for each layer. Graphs represent that mean of convolutional layer in network $CNN\_1C$ reduce to zero at last as compared to other layers while standard deviation (STD) of fc2 layer converges at last. In case of network $CNN\_0C$, mean of fc2 converges at last and STD of fc3 converges at last as compared to other layers.

\section{Conclusion}\label{S5}
This paper is concluded as a low-cost, light weight, simple and easily wearable system is proposed. Laser and high quality webcam is attached it. The system is trained using multicolumn CNN with edges, optical flow and scale space features. These convolutional feature maps are fused involving two kinds of multispectral fusions using addition and maximum. These feature maps are further passed for ROI pooling and classified using two different classification networks $CNN\_0C$ and $CNN\_1C$ having three fully connected layers and one convolutional layer with three fully connected layers respectively. Number of experiments done with these networks show that there is some improvement between $CNN\_0C$ and $CNN\_1C$. However, that is minimum when we are using different kinds of features. Out of all the features mentioned above, scale space features with $CNN\_1C$ outperform the others. The proposed system is designed for helping visually impaired people. It detects and classify obstacles that come on the way and the distance of the obstacle from the user and warns the user about that.

\bibliography{ref_edges-gauss}

\end{document}